%% file: main.tex
\documentclass[letterpaper, 10 pt, conference]{ieeeconf}  

\IEEEoverridecommandlockouts                              

\usepackage{graphics} 
\usepackage{epsfig} 
\usepackage{mathptmx} 
\usepackage{times} 
\usepackage{amsmath} 
\usepackage{amssymb}  
\usepackage{geometry}
\usepackage{color}
\usepackage{algorithm,algpseudocode}
\usepackage{babel}
\usepackage{multirow}
\usepackage{amstext}
\usepackage{units}
\usepackage{booktabs}
\usepackage{caption}
\usepackage{subcaption}
\usepackage{xcolor}
\usepackage{array} 
\usepackage{svg}

\title{\LARGE \bf
DualDiff: Dual-branch Diffusion Model for Autonomous Driving with Semantic Fusion
}

\author{~Haoteng Li$^{*,1}$, Zhao Yang$^{*,1}$, Zezhong Qian$^{1}$, Gongpeng Zhao$^{2}$, \\ ~~~Yuqi Huang$^{1}$, Jun Yu$^{2}$, Huazheng Zhou$^{1}$ and Longjun Liu$^{\dag,1}$
\thanks{* The first two authors contributed equally. This work was supported by the Natural Science Foundation of China under Grant NSFC 62088102.}
\thanks{$\dag$ Corresponding author. Email: {\tt\small liulongjun@xjtu.edu.cn }}
\thanks{$^{1}$ Haoteng Li, Zhao Yang, Zezhong Qian, Yuqi Huang, Huazheng Zhou and Longjun Liu are with National Key Laboratory of Human-Machine Hybrid Augmented Intelligence, National Engineering Research Center for Visual Information and Applications, and Institute of Artificial Intelligence and Robotics, Xi’an Jiaotong University, Xi'an, Shaanxi, 710049, China.}
\thanks{$^{2}$ Gongpeng Zhao and Jun Yu are with University of Science and Technology of China, 
Hefei, Anhui, 230026, China.}
}

\begin{document}

\global\long\def\D{\mathcal{D}}%
\global\long\def\L{\text{L}}%
\global\long\def\th{\boldsymbol{\theta}}%
\global\long\def\seg{\text{seg}}%
\global\long\def\hm{\text{hm}}%
\global\long\def\bbox{\text{bbox}}%
\global\long\def\ri{\text{ri}}%
\global\long\def\fm{\text{fm}}%
\global\long\def\1{\boldsymbol{1}}%
\global\long\def\up{\text{U}_{\text{pixel}}}%
\global\long\def\cp{\text{C}_{\text{pixel}}}%
\global\long\def\po{\text{point}}%
\global\long\def\uh{\text{U}_{\text{hm}}}%
\global\long\def\pse{\text{pseudo}}%
\global\long\def\mar{\text{marginal}}%
\global\long\def\teach{\text{teacher}}%
\global\long\def\score{\text{score}}%
\global\long\def\p{\text{p}}%
\global\long\def\y{\text{y}}%
\global\long\def\h{\text{h}}%
\global\long\def\X{\text{X}}%
\global\long\def\Y{\text{Y}}%
\global\long\def\uc{\text{U}_{\text{class}}}%

\newcommand{\todo}[1]{\textcolor{red}{TODO: #1}}

\maketitle
\thispagestyle{empty}
\pagestyle{empty}

\begin{abstract}
Accurate and high-fidelity driving scene reconstruction relies on fully leveraging scene information as conditioning. However, existing approaches, which primarily use 3D bounding boxes and binary maps for foreground and background control, fall short in capturing the complexity of the scene and integrating multi-modal information. In this paper, we propose DualDiff, a dual-branch conditional diffusion model designed to enhance multi-view driving scene generation. We introduce Occupancy Ray Sampling (ORS), a semantic-rich 3D representation, alongside numerical driving scene representation, for comprehensive foreground and background control. To improve cross-modal information integration, we propose a Semantic Fusion Attention (SFA) mechanism that aligns and fuses features across modalities. Furthermore, we design a foreground-aware masked (FGM) loss to enhance the generation of tiny objects. DualDiff achieves state-of-the-art performance in FID score, as well as consistently better results in downstream BEV segmentation and 3D object detection tasks.
\end{abstract}


    \input{sections/1.intro}

    \input{sections/2.related}
    \input{sections/3.methods}
    \input{sections/4.results}
    \input{sections/5.conclusions}

\input{output.bbl}
\bibliographystyle{IEEEtran}
\addtolength{\textheight}{-12cm}

\end{document}

%% file: sections/1.intro.tex
\section{Introduction}
Most autonomous driving research relies on large-scale camera datasets with detailed annotations \cite{cui2024survey}, \cite{MARTINEZDIAZ2018275}, \cite{chen2024end}. However, due to the high cost of data collection and annotation, open-source vision datasets are limited \cite{BOSCH201876}, \cite{10432820}. Advanced generative models, such as Stable Diffusion \cite{rombach2022high}, \cite{li2023drivingdiffusion}, \cite{song2020denoising}, offer a promising solution by generating realistic images for synthesizing street-view data.
\begin{figure}
\centering{}
\includegraphics[width=0.79\linewidth]{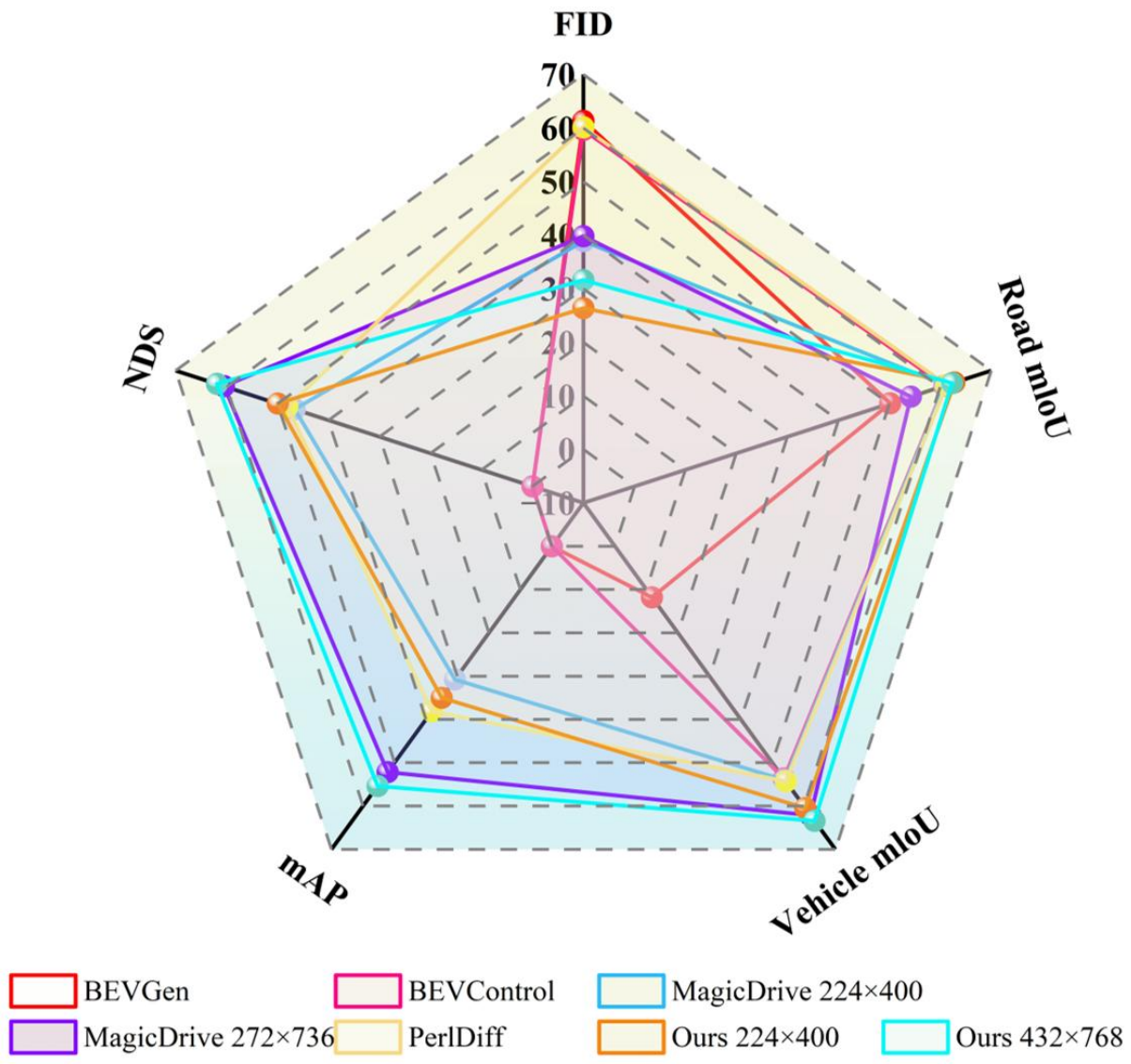}
\caption{We have achieved state-of-the-art performance in several evaluation metrics compared to other custom or base models. To present the data in the charts more clearly, we have scaled some of the metrics.} \label{fig: lidar}
\vspace{-0.8cm}
\end{figure}

Current conditional generation models for autonomous driving have achieved high-fidelity scene generation, aiding downstream visual tasks. However, several limitations persist: 1) \textbf{Limited scene control conditions.} Previous approaches \cite{wen2024panacea}, \cite{gao2023magicdrive}, \cite{zhang2024perldiff}, \cite{yang2023bevcontrol} primarily use 3D bounding boxes for foreground and binary maps for background representation, which are inadequate for capturing the complexity of the driving scene. Moreover, the disparity in modality between sparse 3D boxes and dense binary maps makes it challenging to encode and utilize both information in a unified manner. 2) \textbf{Insufficient cross-modality interaction.} Multi-modal inputs (e.g., maps, bounding boxes, prompts) contain diverse information that needs effective integration for accurate scene generation. Current approaches \cite{li2023drivingdiffusion}, \cite{gao2023magicdrive}, \cite{wang2024drivewm}, \cite{swerdlow2024bevgen}, \cite{wang2023drivedreamer} often rely on simple concatenation, lacking strategies for holistic scene understanding, leading to suboptimal generation outcomes. 3) \textbf{Lack of enhancement of tiny object details.}
Tiny objects are essential for downstream vision tasks, yet current models often neglect their accurate generation, lacking targeted mechanisms to enhance their detail and precision. 

In this paper, we propose DualDiff, a dual-branch architecture with comprehensive scene control for high quality generation. We address the aforementioned issues by the following designs: 1) \textbf{Comprehensive scene control with dual-branch architecture.} We introduce an Occupancy Ray Sampling (ORS) representation, rich in semantic and 3D information. To further supplement ORS with detailed information, we introduce numerical driving scene representation, where we replace the dense binary maps with sparse vectorized maps, aligning with the bounding box modality. We then propose a dual-branch architecture to integrate these representations, achieving unified and balanced foreground-background generation. 2) \textbf{Cross-modality semantic interaction.} To enhance multi-modal input integration, we propose a Semantic Fusion Attention (SFA) mechanism. SFA updates ORS features using multi-modal data from the numerical driving scene representation, to provide fused, aligned scene feature for the generative model. 3) \textbf{Tiny object generation enhancement.} We design foreground-aware masked (FGM) loss by applying a weighted mask to the original denoising loss, a simple yet effective approach to improve the detail generation of distant or tiny objects.

Our proposed model effectively integrates cross-modal scene features, enabling the accurate reconstruction of scene content, and outperforms previous methods in terms of image style fidelity, foreground attributes, and background layout accuracy. The main contributions of this work are summarized below.
\begin{itemize}
\item 
We propose a dual-branch (foreground-background) architecture that leverages our introduced occupancy ray sampling (ORS) representation and numerical driving scene representation, to exert fully control upon the scene reconstruction process.
\item
We propose an efficient semantic fusion attention (SFA) module to improve the understanding of multi-modal scene representations. In addition, we propose a foreground-aware masked (FGM) Loss to further improve the generation of tiny objects.
\item 
Our model surpasses the previous best methods in terms of realistic style reconstruction and accurate generation of foreground and background content, achieving the state-of-the-art (SOTA) performance.
\end{itemize}

%% file: sections/2.related.tex
\section{Related Work}
\noindent{\textbf{Diffusion Models for Conditional Generation.}}
Recently, various methods have been proposed for conditional generation using diffusion models \cite{Parihar_2024_CVPR}, \cite{nakashima2024lidar}. For instance, ControlNet \cite{zhang2023controlnet}, \cite{zhao2024uni} integrates external neural networks to inject conditions, enabling precise control over generated content. Some models \cite{wang2024drivewm}, \cite{wang2023disco} use cross-attention within the UNet architecture, while others \cite{Singer2022makeavideo}, \cite{he2023animate} incorporate conditions by concatenating or element-wise operations with the model's noise. 
However, most of the methods are rooted in other domains, and couldn't directly generalize to driving scene specific types of data, including occupancy voxels, vectorized maps, etc. This paper addresses these challenges by exploring the adaptation of multi-modal conditioning diffusion models with modality specific encoding methods.

\noindent{\textbf{Scene Reconstruction in Automous Driving.}}
With the continuous advancement of autonomous driving, various methods have been proposed for scene reconstruction in autonomous driving scenarios. BEVControl \cite{yang2023bevcontrol} combines bird's-eye view and street view information to generate geometrically consistent foregrounds, while MagicDrive \cite{gao2023magicdrive} integrates BEV maps, 3D bounding boxes, and camera poses, using inter-view attention to capture subtle 3D details. Driving Diffusion \cite{li2023drivingdiffusion}, \cite{zou2024diffbev} emphasize multi-view coherence through the use of 3D layouts derived from map and bounding box, whereas Panacea \cite{wen2024panacea} utilizes BEV sequences to ensure temporal stability. PerlDiff \cite{zhang2024perldiff} employs 3D perspective geometric priors for more precise object control in street view generation. While the above methods rely mainly on modalities of bounding box and binary maps, we seek to utilize comprehensive scene information, including occupancy voxels, bounding boxes and vectorized maps, etc.



%% file: sections/3.methods.tex

\section{Method}
\begin{figure*}
\centering{}
\includegraphics[width=\linewidth]{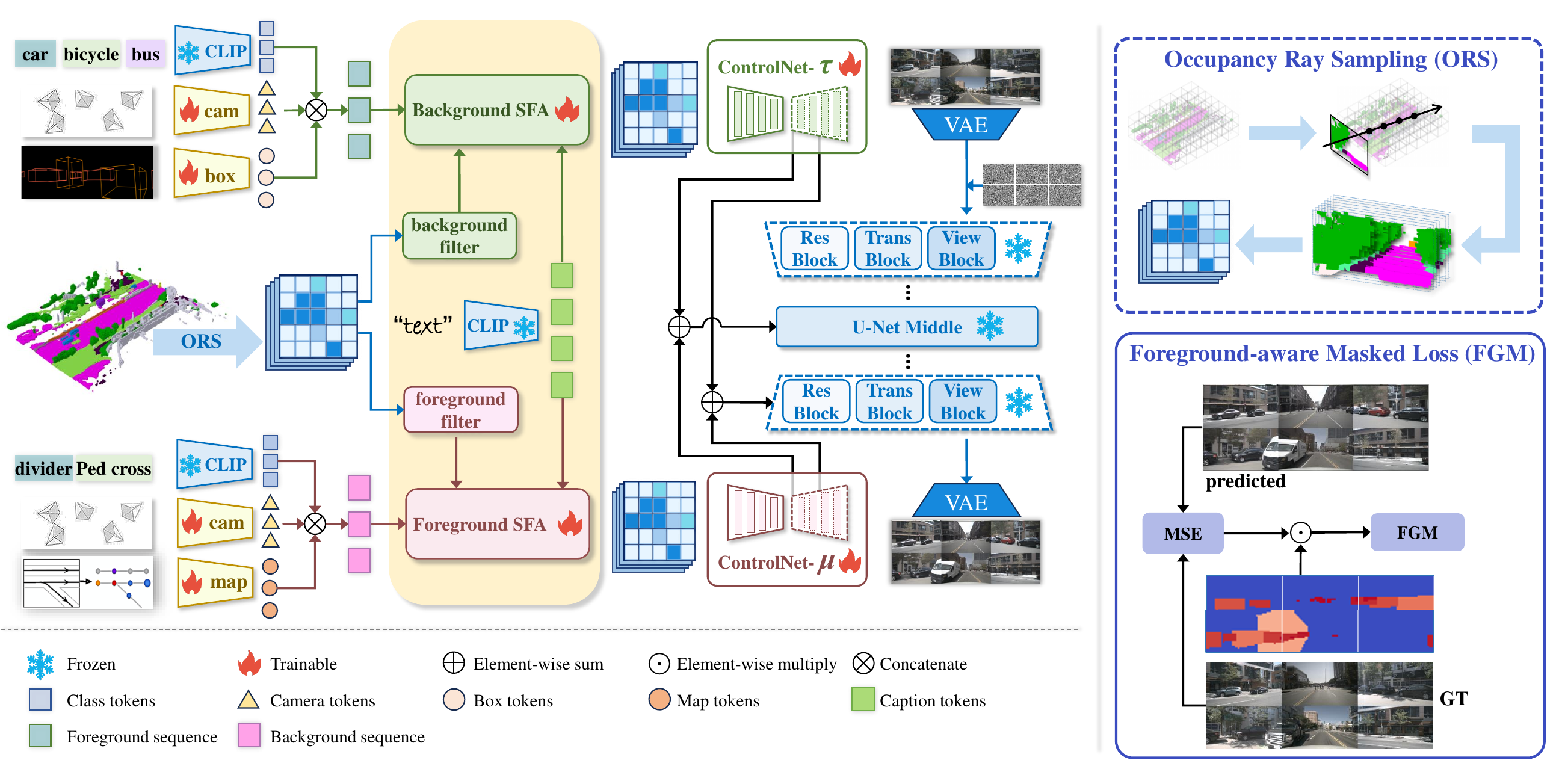}
\caption{Overview of DualDiff for multi-view image generation. We use {occupancy ray sampling (ORS) and numerical driving scene representation}, which are fused through the proposed semantic fusion attention (SFA) module and then used as inputs to the dual-branch foreground-background architecture. The outputs of the branches are then merged back into the UNet in the form of ControlNet residuals to obtain the final output.} \label{fig: main}
\vspace{-0.2cm}
\end{figure*}
{\subsection{Dual-branch Foreground-Background Architecture}
\noindent{\textbf{Occupancy Ray Sampling Representation.}}
To effectively leverage the 3D layout and semantic information from the occupancy voxels, we propose an occupancy ray sampling (ORS) strategy, that captures condensed feature from the raw occ voxels. The ORS method can be analogized to the physical imaging process, where rays emanating from objects converge onto the imaging plane to form an image, while we actively emit rays from the camera to detect occupied voxels. For each pixel $s_\text{img}$ on the imaging plane with size $U \times V$, a ray is emitted in the direction $r$, along which we sample equidistant points $\hat{s}_{\text{ego}}$. Each point in $\hat{s}_{\text{ego}}$ serves as a query index to the raw voxels $O \in \mathbb{R}^{H \times W \times D}$ and records the occupancy state at its position, leading to the condensed feature $v \in \mathbb{R}^{U \times V \times N}$. The whole process is as follows:
\begin{equation}
\begin{gathered}
r = \text{norm}(T^{-1} \cdot K^{-1} \cdot s_{\text{img}} - p_{\text{ego}}) \\
\hat{s}_{\text{ego}} = \{p_{\text{ego}} + r \cdot n\mid n \in N\}\\
v = \mathcal{F}_\text{ORS}(O, \hat{s}_{\text{ego}})
\end{gathered}
\end{equation}
where $\mathcal{F}_\text{ORS}(\cdot)$ represent the ORS function. $K$ and $T$ denote the camera's intrinsic and extrinsic parameters. $N$ represents the set of ray sampling step sizes, and $p_{\text{ego}}$ specifies the camera's position in the ego vehicle's coordinate system.

\noindent{\textbf{Numerical Driving Scene Representation.}} 
In addition to the global structure and semantic information captured by ORS, we also provide fine-grained details to the model by introducing a digital scene representation, which also alleviates the detail loss problem of features during UNet downsampling.
Specifically, we use 3D bounding boxes to represent foreground objects, vectorized maps to represent background details, and camera poses and textual cues to control abstract context information. First, for the foreground object $B = \{(t^b_i, b_i)\}_{i=1}^{N}$, $b_i = \{(x_j, y_j, z_j)\}_{j=1}^{8} \in \mathbb{R}^{8 \times 3}$ represents the 3D bounding box coordinates, and $t^b \in \mathcal{T}_\text{box}$ represents the object category (consistent with \cite{gao2023magicdrive}). For a vectorized map $M = \{(t^m_i, m_i)\}_{i=1}^{N}$, $m_i = \{(v_j)\}_{j=1}^{8} \in \mathbb{R}^{8 \times 3}$ represents an ordered set of points of map elements with category $t^m \in \mathcal{T}_\text{map}$ (pedestrian crossing, divider and boundary as defined in \cite{MapTR}). Then, we use CLIP \cite{radford2021clip} to encode the category information of the bounding box and map elements, and input the corresponding 3D coordinate information into Fourier embedding \cite{mildenhall2021nerf}. The final box and map features obtained by splicing are as follows:
\begin{gather}
\quad c_\text{box} = E_{\text{box}}([\mathrm{CLIP}(t^b), \mathrm{Fourier}(b)]) \\ 
\quad c_\text{map} = E_{\text{map}}([\mathrm{CLIP}(t^m), \mathrm{Fourier}(m)]) 
\end{gather}

We also provide the model with camera poses for view-specific generation, and textual hints for abstract semantic control descriptions. {For textual hints, We use a frozen CLIP as the feature extractor}. For camera poses, we concatenate intrinsic parameters, rotation parameters, and translation parameters before feature extraction. The detailed process is as follows:
\begin{gather}
c_{\text{text}} = E_{\text{text}}(\mathrm{CLIP}(L))\\
\quad c_{\text{cam}} = E_{\text{cam}}(\mathrm{Fourier}([K, R, t]^T))
\end{gather}
where $L$ refers to text prompts, and $K$, $R$, $t$ correspond to the intrinsic parameters, rotation, and translation of the camera.

\noindent{\textbf{{Dual-branch Architecture.}}
We propose a dual-branch conditional control structure using the previously introduced driving scene representations. The overall design is depicted in Figure \ref{fig: main}. For the background branch $\tau$, we filter out foreground grids based on the semantic labels from the occupancy grids and perform ORS to obtain condition $v_b$. We then concatenate features of camera poses, textual information and numerical foreground bounding boxes, yielding $c_{\text{env}} = 
[c_{\text{cam}}, c_{\text{text}},c_\text{box}]
$ as the input to the cross-attention module of $\tau$. {The process is the same for the foreground branch $\mu$, where we provide filtered foreground ORS feature $v_f$, and $c_{\text{env}} = 
[c_{\text{cam}}, c_{\text{text}},c_\text{map}]
$}. To this end, we have completed the construction of the dual-branch architecture, where the inputs processed by the two branches exhibit a dual relationship.
\begin{figure}
\centering{}
\includegraphics[width=1.0\linewidth]{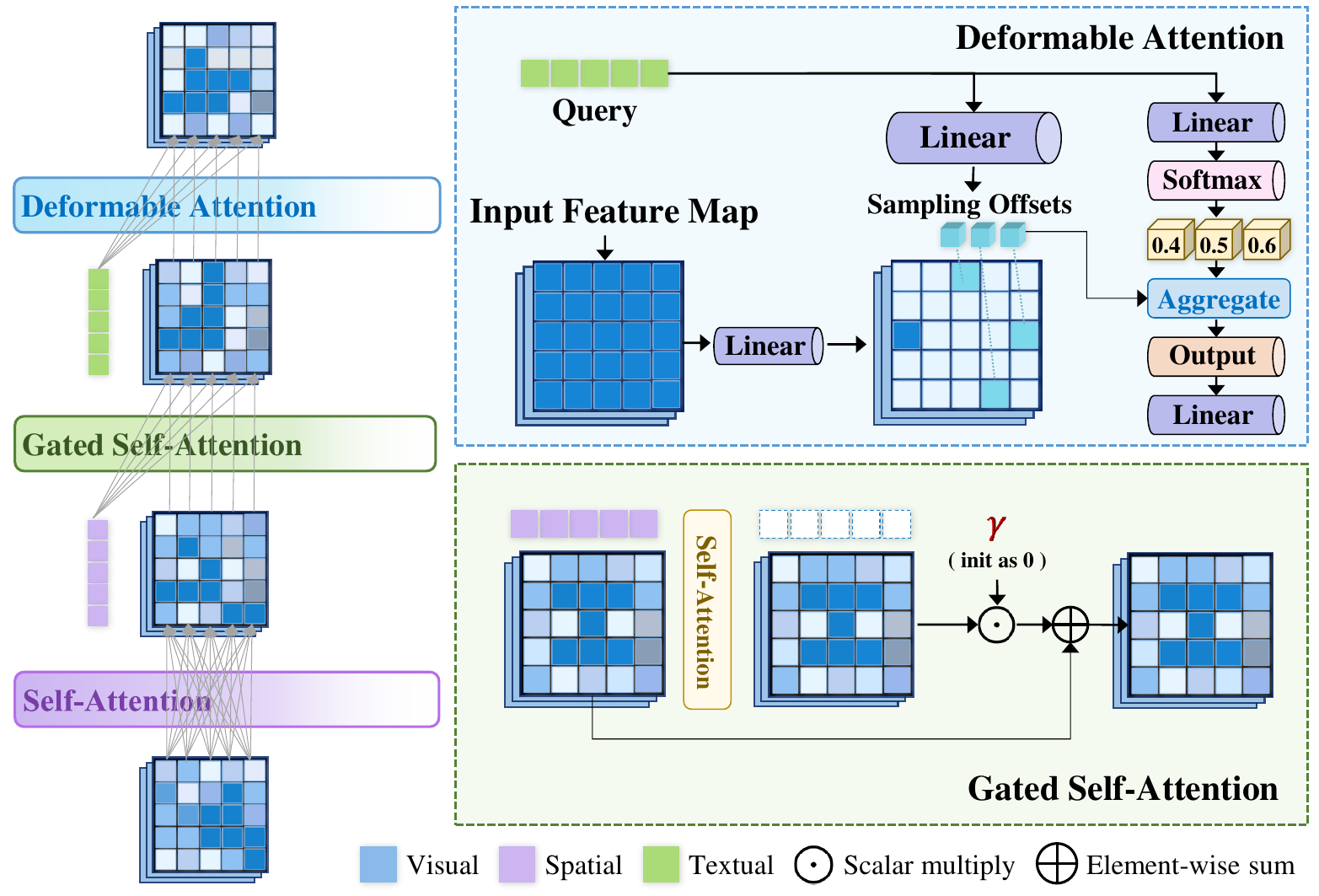}
\caption{Illustrations of our proposed Semantic Fusion Attention (SFA), which sequentially fuses ORS features with multi-modal information.} \label{fig: fusion}
\vspace{-0.4cm}
\end{figure}
\subsection{Multi-modal Scene Representations Alignment}
In autonomous driving scene generation, various modalities can be used as control inputs to the model, such as visual projected bounding boxes or map masks, vehicle bounding box coordinates or vector maps, and text descriptions. A single visual modality often fails to capture the comprehensive details required for scene generation. Here, we use the previously introduced ORS feature $v$ as the visual modality input because of its camera-like sampling properties, {the numerical feature $c_\text{box}$ and $c_\text{map}$} as the spatial modality, and $c_\text{text}$ as the semantic text modality. Based on the multi-modal input, we design semantic fusion attention (SFA) to sequentially update the initial visual feature $v$ with multi-modal information, thereby using the fused and updated feature $v^*$ as the input of the dual branches.

The structure of the SFA module is shown in Figure \ref{fig: fusion}. We first apply self-attention to the ORS visual feature \(v\), and then use the gated self-attention mechanism on the concatenation of the visual and spatial features $[v^{\prime}_1, c_\text{spatial}]$ to achieve spatial grounding, obtaining the updated visual feature \(v^{\prime}_2\). Finally, the updated feature $v^{\prime}_2$ (rich in spatial layout information) is fused with the textual features using the deformable attention mechanism to obtain the output $v^*$.
\begin{gather}
v^{\prime}_1 = v + \text{SelfAttn}\left(v\right) \\ 
v^{\prime}_2 = v^{\prime}_1 + \tanh(\gamma) \cdot \text{SelfAttn}\left([v^{\prime}_1, c_\text{spatial}]\right) 
\\
v^* = \text{DeformAttn}(v^{\prime}_2, \boldsymbol{c}_{\text{text}}) 
\end{gather}
where \( \gamma \) is a learnable scalar (initialized to 0). The variable $c_\text{spatial}$ refers to $c_\text{map}$ at the foreground branch $\mu$, and $c_\text{box}$ at the background branch $\tau$. We carry out the calculation of SFA for $v_f$ and $v_b$ seperately on the front of each branch.

By integrating multiple modalities such as vision, space, and text, the model can effectively capture the complex semantics and dynamics of autonomous driving scenes, thereby producing more realistic, contextually accurate, and geometrically consistent outputs. SFA improves model performance with minimal additional trainable parameters, particularly in FID score, as shown in the experiments.
\subsection{Conditional Diffusion Model with Foreground Mask}
Different from the previous diffusion model denoising loss function method, in order to enhance the model's ability to generate distant and small objects, we proposed the foreground-aware masked loss (FGM), which adjusts the weight of the denoising loss according to the size of the foreground objects in the image plane. Experimental results show that FGM Loss can effectively improve the quality of foreground generation with only a simple modification to the original loss. We use the camera-view projected bounding boxes of the foreground objects to construct a loss mask $m$, where we assign higher values to the area of smaller boxes, specified as follows:
\begin{equation}
    {{m}_{ij}} = \left\{
    \begin{array}{ll}
    2 - \frac{a_{ij}}{U \times V} & (i,j) \in \text{foreground}, \\
    1 & (i,j) \in \text{background}
    \end{array}
    \right.
\end{equation}
where $a_{ij}$ represents the area of the foreground mask at coordinate $(i,j)$, and $U$, $V$ represent the width and height of the noise image features. The network is trained to predict the noise by minimizing the mean square error:
\begin{gather}
\begin{split}
\min _\theta \mathcal{L} = &\mathbb{E}_{\mathcal{E}(\boldsymbol{x}_0), c_\text{env}, \tau_\theta(v^*_b), \mu_\theta(v^*_f), \epsilon \sim \mathcal{N}(0,1), t} \\ 
&\left[ \|\epsilon - \epsilon_{\theta}(z_t, t, c_\text{env}, \tau_\theta(v^*_b), \mu_\theta(v^*_f))\|_{2}^{2} \right] \odot m
\end{split}
\end{gather}
{where $z_0=\mathcal{E}(\boldsymbol{x}_0)$ is the hidden feature of the original image $x_0$ in the space of autoencoder. $z_0$ is diffused $t$ time steps to produce noisy latents $z_t$. $\tau_\theta$ and }$\mu_\theta$ refer to trainable dual-branch and $\epsilon_\theta$ is frozen. $v_b^*$, $v_f^*$ are ORS feature updated by SFA and $c_\text{env}$ refers to concatenated numerical scene representations. 

%% file: sections/4.results.tex
\section{Experiments}
\input{tables/val_table}
\subsection{Experimental Setups}
\noindent{\textbf{Dataset}.} We train our model on the open-source nuScenes \cite{caesar2020nuscenes} dataset of 1000 driving scenes collected across Singapore and Boston, with annotations including 3D foreground, maps, and supplementary occupancy annotations \cite{tian2023occ3d} for each driving scene. Our model is trained on 750 training scenes and evaluated on 150 validation scenes. We also extend our model to the Waymo \cite{Sun_2020_waymo} dataset. To match the amount of training data in the nuScenes task, we use 150 of the 798 scenes in the Waymo training set and evaluate on all 202 validation scenes.

\noindent{\textbf{Evaluation Metrics.}}
Following previous methods, we evaluate the overall fidelity of the generated image style as well as the accuracy of the foreground and background in the images. We use Fréchet Inception Distance (FID) to assess the realism of the image style. For evaluating the accuracy of driving scene generation, we use CVT \cite{zhou2022cvt} and BEVFusion \cite{liu2022bevfusion} on the nuScenes task by directly testing the generated validation set with pre-trained models. In the task of supporting model training using the generated training set, we report results on StreamPETR \cite{wang2023streampetr}. For the Waymo task, we adopt BevFormer \cite{li2022bevformer}.

\noindent{\textbf{Implementation Details}}
We initialize our model using Stable Diffusion v1.5 and the corresponding version of ControlNet pretrained for segmentation tasks, keeping the UNet frozen throughout training. We train the dual foreground and background branches separately for 80 epochs, followed by combined training of the entire dual-branch model for 30 epochs, with the learning rate set to $8\times10^{-5}$. For inference, we use UniPC \cite{zhao2024unipc} for 20 steps of sampling, with CFG set to 2. Resolution is set to 224$\times$400 for the nuScenes task, and 320$\times$480 for the Waymo task.

\subsection{Main Results}
\noindent{\textbf{Scene Reconstruction on nuScenes.}}
In this section, we report the image style quality and accuracy of content on the nuScenes dataset. Our model surpasses all previous and concurrent methods in its ability to reconstruct the style of real street scenes, achieving the lowest FID score of 10.99 as shown in Table \ref{tab:fid}. In terms of content accuracy in Table \ref{tab:main}, we further raise the upper limits of BEV segmentation tasks and 3D object detection tasks. Notably, we also report our results at a resolution of 432$\times$768, for a fair comparison with models at higher resolutions.

\input{tables/kitti}
\noindent{\textbf{Scene Reconstruction on Waymo.}}
We further validate the effectiveness of our proposed method on the Waymo dataset. Since no previous methods have reported performance metrics on this dataset, we adapt the open-source model MagicDrive to Waymo, as the baseline for comparison in this task. Notably, for this task, we only use the background branch of our proposed model and keep the training rounds consistent with the baseline. On the one hand, we demonstrate the performance improvements brought by introducing the occupancy and fusion modules in Table \ref{table:bev_detection_Waymo} while keeping the amount of learnable parameters comparable. On the other hand, we illustrate the further improvements that can be achieved by the dual-branch setup compared to the single-branch setup in the ablation studies.

\noindent{\textbf{Training Support for Downstream Perception Tasks.}}
The training support experiment aims at evaluating the improvements on the performance of downstream vision-based perception models brought by diverse generated training data. We use our model to generate a simulated dataset and trained the perception model StreamPETR \cite{wang2023streampetr} on a mix of real and simulated data. The results in Table \ref{tab:streampetr_train} show that our method can support the training of vision-based perception model with consistently better results across all the listed metrics.

\begin{table}
\centering{}%
\scalebox{0.96}{
\begin{tabular}{c|c|c|c|c}
\toprule
Data & mAP $\uparrow$ & mAOE $\downarrow$ & mAVE $\downarrow$ & NDS $\uparrow$ \\
\midrule

   w/o synthetic data & 34.5 & 59.4 & 29.1 & 46.9 \\


   w/ DualDiff
   & \textbf{36.2 {\tiny\textcolor{green}{+1.7\%}}}
   & \textbf{46.6 {\tiny\textcolor{green}{-12.8\%}}}
   & \textbf{27.9 {\tiny\textcolor{green}{-1.2\%}}}
   & \textbf{49.6 {\tiny\textcolor{green}{+2.7\%}}}
   \\
\bottomrule
\end{tabular}
}
\caption{Comparison about support training for 3D object detection model (StreamPETR \cite{wang2023streampetr}). Results are reported on the nuScenes validation set.}
\vspace{-0.5cm}
\label{tab:streampetr_train}
\end{table}

\subsection{Ablation Study}\label{sec:ablation}
\begin{figure}
\centering{} 
\includegraphics[width=0.98\columnwidth]{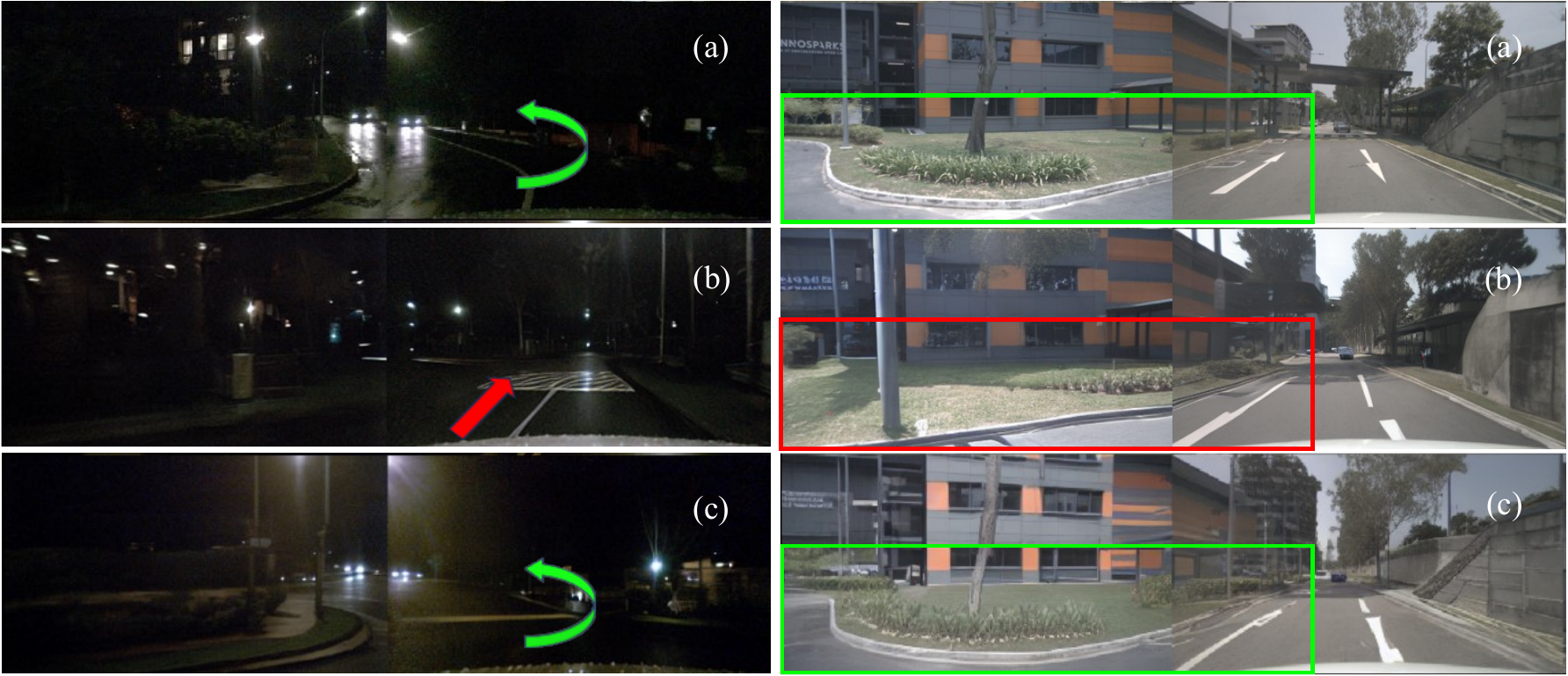}
\caption{Driving scenes of (a) ground truth, (b) MagicDrive and (c) DualDiff (Ours). Compared to the baseline, DualDiff faithfully reproduces the left turn orientation as well as the car in distance in the night scene, while in the daylight case, DualDiff generates the edge of the road as well as the tree
behind precisely.}
\vspace{-0.3cm}
\label{fig: road night case}
\end{figure}

\begin{figure}
\centering{}  
\includegraphics[width=0.95\columnwidth]{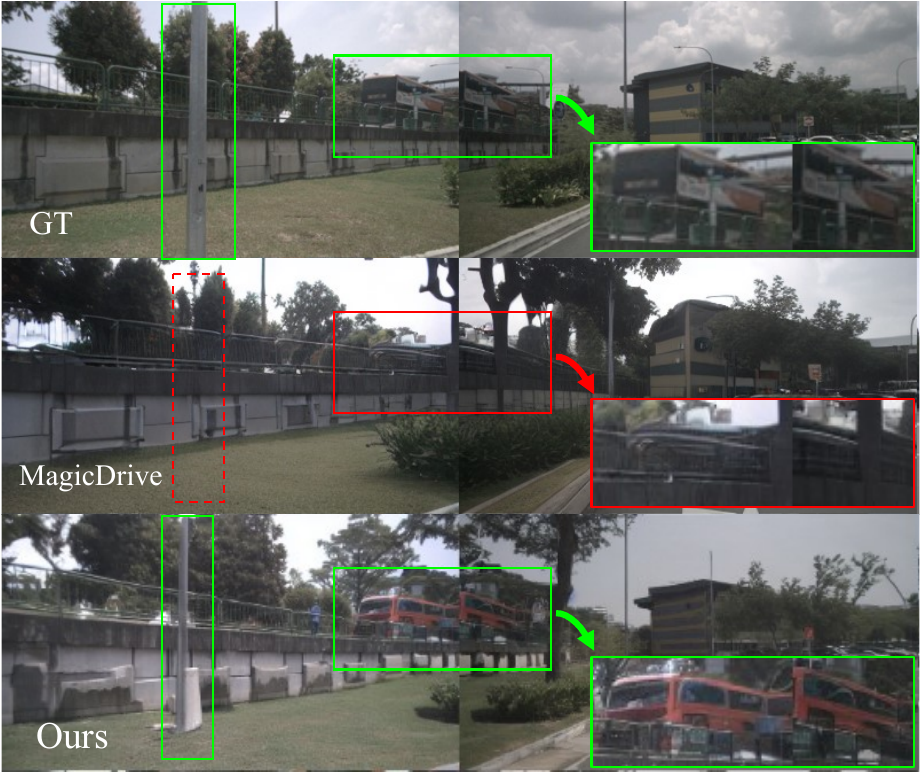}
\vspace{-0.1cm}
\caption{Reconstruction scene in daylight, where our model generates the bus in distance and the lamp pole correctly.}
\vspace{-0.7cm}
\label{fig: day road case}
\end{figure}

\noindent{\textbf{Occupancy-based Representations.}}
We construct our model based on the open-source MagicDrive. Compared to the baseline, we replace the BEV map condition with the occ-based ORS feature. As illustrated in the figure \ref{fig: road night case}, our model generates correct road layouts and edge details. Compared to the BEV map, the ORS feature maintains viewpoint consistency with the ground truth street-view images, which facilitates faster convergence in ControlNet. The significant improvement in Road mIoU, as shown in Table \ref{tab:main} further supports our conclusion. Moreover, beyond the advantage of viewpoint projection, the ORS feature provides details that are missing from foreground object annotations. As depicted in the figure \ref{fig: day road case}, our model accurately reconstructs the lamp pole which is not annotated as ordinary categories in the dataset.

\begin{figure}
\centering{}  
\includegraphics[width=0.86\columnwidth]{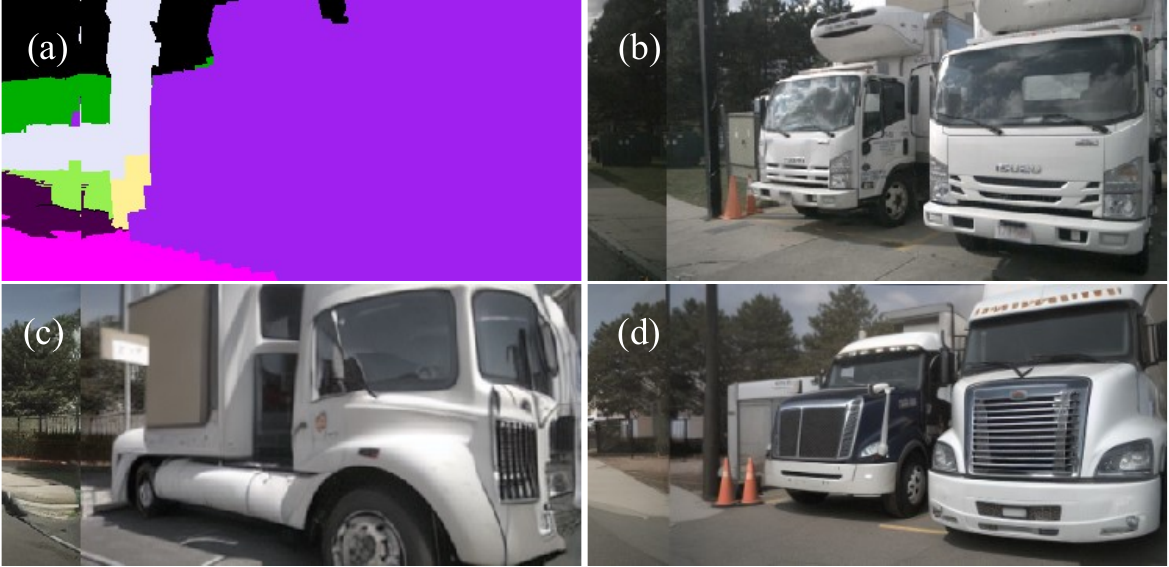}
\caption{Reconstruction of two trucks. Presented in the figures are (a) occ visualization, (b) ground truth, (c) ours without numerical bounding box represenations, (d) ours.}
\vspace{-0.3cm}
\label{fig: trucks}
\end{figure}
\noindent{\textbf{Numerical Representations.}} Numerical representations encompass both foreground and background elements. The bounding box (bbox) offers supplementary information for the ORS features. As shown in the figure \ref{fig: trucks}, two trucks that are so close to each other that the occ of the trucks are annotated as a whole. In this case, the introduction of bbox information allows us to successfully reconstruct the trucks. Additionally, the inclusion of map information reinforces road elements in the generated scenes. 
\begin{table}
\centering{}%
\scalebox{0.84}{
\begin{tabular}{ccc|ccc}
\toprule
Dual Branch & SFA & FGM Loss & FID$\downarrow$ & Road mIoU$\uparrow$ & Vehicle mIoU$\uparrow$ \\
\midrule
& & & 13.26 & 62.19 & 27.30 \\
& & \checkmark & 12.95 & 61.78 & 29.00 \\
&  \checkmark & \checkmark & 12.57 & 61.47 & 29.11 \\
\checkmark & \checkmark & & {11.01} & {62.71} & {29.19} \\
\checkmark & \checkmark & \checkmark & \textbf{10.99} & \textbf{62.75} & \textbf{30.22} \\
\bottomrule
\end{tabular}
}
\vspace{-0.1cm}
\caption{Ablation studies on proposed modules with respect to FID score and BEV segmentation task (CVT).}
\vspace{-0.6cm}
\label{tab:ablation}
\end{table}

\noindent{\textbf{Quantitative Analysis on Proposed Modules.}} 
In Table \ref{tab:ablation}, we demonstrate the effectiveness of each proposed design. Notably, in the configuration without the dual-branch setup, we use the background ControlNet branch. Thanks to the foreground-aware loss enhancement, which focuses on the model's predictions in the foreground areas, particularly for tiny objects, we achieve a significant improvement in foreground-related vehicle mIoU in both single and dual-branch settings. The proposed Semantic Fusion Attention effectively aligns multi-modal input information, enhancing the model's overall understanding of the scene to be reconstructed, thereby further reducing the FID score. Finally, compared to using only the background ControlNet, the introduction of the dual-branch setup adds the necessary modules for foreground generation and efficiently integrates all input information, leading to significant improvements in all metrics, achieving state-of-the-art performance.


%% file: tables/val_table.tex
\begin{table*}
\vspace{-0.8cm}
\centering{}%
\scalebox{1.15}{
\begin{tabular}{l|c|c|c|c|c|c|c}
\toprule
\multirow{2}[3]{*}{Method} &
  \multirow{2}[3]{*}{\begin{tabular}[c]{@{}c@{}}Synthesis\\ resolution\end{tabular}} &
  \multirow{2}[3]{*}{FID$\downarrow$} &
  \multicolumn{2}{c|}{BEV segmentation} &
  \multicolumn{2}{c|}{3D object detection} & 
  \multirow{2}[3]{*}{Reference}\\
  \cmidrule{4-7} 
  &&&
  Road mIoU $\uparrow$ &
  Vehicle mIoU $\uparrow$ &
  mAP $\uparrow$ &
  NDS $\uparrow$   \\
\midrule
Oracle\cite{gao2023magicdrive}
    & -
    & -
    & 72.21
    & 33.66
    & 35.54
    & 41.21 
    & ICLR\textcolor{blue}{2024}\\
Oracle\cite{gao2023magicdrive}
    & 224$\times$400
    & -
    & 72.19
    & 33.61
    & 23.54
    & 31.08 
    & ICLR\textcolor{blue}{2024}\\
\midrule
BEVGen\cite{swerdlow2024bevgen}
    & 224$\times$400
    & 25.54
    & 50.20
    & 5.89
    & -
    & -     
    & RAL\textcolor{blue}{2024} \\
BEVControl\cite{yang2023bevcontrol}
    & -
    & 24.85
    & 60.80
    & 26.80
    & -
    & -    
    & Arxiv\textcolor{blue}{2023} \\ 
MagicDrive\cite{gao2023magicdrive}
    & 224$\times$400
    & 16.20
    & 61.05
    & 27.01
    & 12.30
    & 23.32 
    & ICLR\textcolor{blue}{2024} \\
MagicDrive\cite{gao2023magicdrive}
    & 272$\times$736
    & 16.59
    & 54.24
    & 31.05
    & 20.85
    & 30.26 
    & ICLR\textcolor{blue}{2024} \\
PerlDiff\cite{zhang2024perldiff}
    & 256$\times$704
    & 25.06
    & 61.26
    & 27.13
    & 15.24
    & 24.05 
    & Arxiv\textcolor{blue}{2024} \\
\hline
\textbf{DualDiff} (\textbf{Ours})
    & 224$\times$400
    & \textbf{10.99}
    & \textbf{62.75}
    & 30.22
    & 13.99
    & 24.98
    & - \\
\textbf{DualDiff} (\textbf{Ours})
    & 432$\times$768
    & 13.16
    & 62.38
    & \textbf{31.69}
    & \textbf{22.13}
    & \textbf{30.96}
    & - \\
\bottomrule
\end{tabular}
}
\caption{Comparison of generation fidelity among various driving-view generation methods. The synthesis conditions are derived from the nuScenes validation set, and each task employs models trained on the corresponding nuScenes training set. DualDiff consistently outperforms all baseline models across the evaluation metrics.} \label{tab:main}
\vspace{-0.2cm}
\end{table*}




\begin{table*}[t!]
\label{tab:fvd}
\vspace{-1mm}
\begin{center}
\scalebox{0.75}{
\begin{tabular}{c|ccccccccccc}
\toprule
\multirow{2}{*}{\textbf{Metric}} 
& WoVoGen\cite{lu2023wovogen}
& BEVGen\cite{swerdlow2024bevgen} 
& PerlDiff\cite{zhang2024perldiff}
& DriveDreamer-2\cite{zhao2024drivedreamer2}
& BEVControl\cite{yang2023bevcontrol}
& Panacea\cite{wen2024panacea} 
& MagicDrive\cite{gao2023magicdrive} 
& SimGen\cite{zhou2024simgen}
& Delphi\cite{ma2024delphi}
& Drive-WM\cite{wang2024drivewm}

& \textbf{DualDiff}\\

& ECCV\textcolor{blue}{2024}
& RAL\textcolor{blue}{2024} 
& Arxiv\textcolor{blue}{2024}
& Arxiv\textcolor{blue}{2024}
& Arxiv\textcolor{blue}{2023}
& CVPR\textcolor{blue}{2024}
& ICLR\textcolor{blue}{2024} 
& Arxiv\textcolor{blue}{2024}
& Arxiv\textcolor{blue}{2024}
& CVPR\textcolor{blue}{2024}
 
& (\textbf{Ours}) \\
\midrule
FID$\downarrow$ 
& 27.6 
& 25.54 
& 25.06 
& 25.0 
& 24.85 
& 16.96 
& 16.20 
& 15.6 
& 15.08 
& 12.99 
 
& \textbf{10.99}\\
\bottomrule
\end{tabular}
}
\end{center}
\vspace{-2mm}
\caption{Comprehensive comparison of generation fidelity across previous methods trained with nuScenes. DualDiff outperforms all the previous and concurrent street scene reconstruction models on nuScenes validation set.}
\vspace{-4mm}
\label{tab:fid}
\end{table*}

%% file: tables/kitti.tex
\begin{table}
\centering{}%
\setlength{\tabcolsep}{12pt}
\scalebox{0.95}{
\begin{tabular}{c|c|c|c}
\toprule
Metric & Oracle & MagicDrive* & \textbf{DualDiff(Ours)} \\
\midrule
AP $\uparrow$ & 24.90 & 3.10 & \textbf{10.40} \\
FID $\downarrow$ & - & 17.16 & \textbf{11.45} \\




\bottomrule
\end{tabular}
}
\caption{For Waymo task, we report the detection results with BEVFormer \cite{li2022bevformer} and the FID score. For both methods, we use nuScenes pretrained weights as initialization.}
\vspace{-0.5cm}
\label{table:bev_detection_Waymo}
\end{table}

%% file: sections/5.conclusions.tex
\section{Conclusion}
This paper presents DualDiff, a dual-branch foreground-background architecture for conditional driving scene generation. Our model, taking Occupancy Ray Sampling representation and numerical driving scene representation as inputs, with cross-modal information alignment brought by Semantic Fusion Attention, is capable of establishing better understanding of the whole driving scenario. Besides, we design a Foreground-aware Masked loss, a simple modification to the original denoising loss that effectively increase the  performance in tiny object generation. Experiments show that our model establishes new state-of-the-art in both style fidelity and content accuracy.